%% file: main.tex
\documentclass{article}

\usepackage[preprint]{neurips_no_mention}

\usepackage[utf8]{inputenc}
\usepackage[T1]{fontenc}
\usepackage{hyperref}
\usepackage{url}
\usepackage{booktabs}
\usepackage{amsfonts}
\usepackage{amsmath}
\usepackage{amssymb}
\usepackage{nicefrac}
\usepackage{microtype}
\usepackage{graphicx}
\usepackage{xcolor}
\usepackage{subcaption}
\usepackage{multirow}
\usepackage{enumitem}  

\newcommand{\cascade}{\textsc{Dual-Stream}}

\title{Engineering Verifiable Modularity in Transformers via Per-Layer Supervision}

\author{
  Clayton Kerce  \\
  Georgia Tech Research Institute \\
  \texttt{clayton.kerce@gtri.gatech.edu}
}

\begin{document}

\maketitle

\input{sections/00_abstract}
\input{sections/01_intro}
\input{sections/02_features}
\input{sections/03_architecture}
\input{sections/04_validation}
\input{sections/05_causal}
\input{sections/06_discussion}
\input{sections/07_related}
\input{sections/08_conclusion}

\bibliographystyle{plainnat}

\input{main.bbl}
\appendix
\input{sections/appendix_architecture}

\input{sections/appendix_feature_specification}

\input{sections/appendix_topk_clustering}

\end{document}

%% file: sections/00_abstract.tex
\begin{abstract}
Transformers resist surgical control. Ablating an attention head identified as critical for capitalization produces minimal behavioral change because distributed redundancy compensates for damage. This Hydra effect renders interpretability illusory: we may identify components through correlation, but cannot predict or control their causal role.

We demonstrate that architectural interventions can expose hidden modularity. Our approach combines dual-stream processing separating token and contextual representations, per-layer supervision providing independent gradient signal at each depth, and gated attention regularizing toward discrete activation patterns. When trained with per-layer supervision, models produce ablation effects 5 to 23 times larger than architecturally identical controls trained with standard objectives. This enables 4 times greater control leverage on targeted behaviors: scaling identified attention heads produces smooth, predictable changes in model output.

The key finding is architectural. Without per-layer supervision, ablation damage concentrates near zero with low variance (Winograd standard deviation 0.63\%). With per-layer supervision, effects spread widely (standard deviation 6.32\%), revealing which predictions depend on which circuits. The larger variance is not measurement noise but the signature of unmasked modularity.

We validate our approach through three components: engineered features that capture computational dynamics rather than vocabulary structure (validated by near-zero correlation with raw activation clustering), an architecture providing positive control for modularity, and causal experiments demonstrating functional reorganization where different tasks route through different attention heads. This establishes a methodology for transforming interpretability from passive observation to active control. \footnote{This work was partially supported by DARPA contract HR001125C0302.}
\end{abstract}

%% file: sections/01_intro.tex
\section{Introduction}
\label{sec:intro}

Current interpretability methods identify components through correlation but cannot predict their causal role. Attention visualizations show where models allocate weight, but attention correlates weakly with feature importance \citep{jain2019attention, wiegreffe2019attention}. Probing classifiers demonstrate that representations encode linguistic structure, but cannot determine whether ablating a component will change behavior \citep{belinkov2022probing}. These approaches observe computation without testing whether identified components are necessary for the behaviors they appear to implement.

The barrier is structural. Standard transformers exhibit what \citet{mcgrath2023hydra} termed the Hydra effect: ablating an attention head produces minimal behavioral change because distributed redundancy compensates for damage. \citet{michel2019sixteen} demonstrated that most attention heads can be removed at test time with negligible performance degradation. \citet{voita2019analyzing} found that only a few specialized heads perform critical functions while the rest serve as backup systems. The network maintains multiple redundant pathways for each computation, rendering surgical intervention ineffective.

This redundancy is not accidental. The residual stream functions as a shared communication channel \citep{elhage2021mathematical}, enabling virtual weights that distribute computation across components. When one pathway is damaged, others compensate through the residual connection. The result is a system where every component appears somewhat important in isolation but none are individually necessary. We can identify circuits through careful analysis \citep{olsson2022context, wang2023interpretability}, but these circuits prove fragile under intervention because the model routes around damage.

We demonstrate that architectural constraints can force computation into verifiable modular pathways. Our approach combines three mechanisms. First, dual-stream processing separates discrete token identity from accumulated contextual representations, limiting the channels through which compensation can occur. Second, per-layer supervision provides independent gradient signal at each depth, preventing the lazy accumulation of redundant backup circuits that standard training permits. Third, gated attention learns query-dependent head activation, regularizing toward patterns where heads specialize rather than contributing small amounts everywhere.

The key comparison is architectural. We train two models with identical structure: both use dual-stream processing with frozen symbolic streams and gated attention. One receives gradient signal only from the final layer, matching standard transformer training. The other receives per-layer supervision at every depth. If per-layer supervision merely adds computational overhead, both models should show similar ablation sensitivity. If it exposes genuine modularity, the supervised model should exhibit larger, more predictable effects.

The results validate the hypothesis. Control models absorb ablation damage with tight variance around zero (Winograd ablation effect: mean 0.05\%, standard deviation 0.63\%). Models trained with per-layer supervision show wide variance (mean 1.15\%, standard deviation 6.32\%), revealing which predictions depend on which circuits. The supervised model provides 4 times greater control leverage on capitalization behavior: scaling identified attention heads from 0 to 1.5 times baseline produces smooth, monotonic changes in output probability.

We make four contributions. First, we establish a feature engineering methodology based on three invariances that captures computational dynamics rather than vocabulary structure. Our features cluster raw activations with adjusted Rand index 0.008, confirming they discover structure orthogonal to token identity. Second, we demonstrate that per-layer supervision converts silent interference into exposed sensitivity, producing ablation effects 5 to 23 times larger than control models. Third, we show functional reorganization: different tasks route through different attention heads in the supervised model but show entangled routing in controls. Fourth, we provide causal validation through targeted ablation and steering experiments that predict behavioral changes from feature analysis.

The approach establishes that modularity can be engineered rather than merely discovered. Standard training produces redundant pathways because gradient-based optimization encounters no pressure toward interpretability. Architectural constraints and training objectives that explicitly reward independent layer contributions force computation into verifiable circuits. We do not claim this methodology scales immediately to web-scale models or generalizes beyond the architectural family we study. We demonstrate foundational methodology for verifying that identified computational modes have causal consequences, transforming interpretability from passive observation to active control.

%% file: sections/02_features.tex
\section{Feature Engineering for Mode Discovery}
\label{sec:features}

Clustering raw activations rediscovers vocabulary structure. The residual stream is dominated by token embeddings, placing semantically similar words near each other in representation space \citep{elhage2021mathematical}. Sparse autoencoders decompose these representations into interpretable features \citep{bricken2023monosemanticity}, but these features encode what information is present, not how the model processes it. To discover computational modes rather than semantic categories, we extract features that capture relational structure: which attention heads activate, how predictions evolve across layers, how attention distributes over context.

We impose three design constraints that ensure features remain independent of vocabulary and position while preserving the structure needed for manifold learning.

\subsection{Design Principles}
\label{sec:design-principles}

\paragraph{Token Identity Invariance.}
Features capture how the transformer processes tokens, not which tokens appear. Two predictions with identical layer-wise probability trajectories and attention patterns produce similar feature vectors whether the context contains ``cat'' or ``dog.'' This constraint isolates computational strategy from semantic content.

\paragraph{Permutation Invariance.}
Features cannot depend on absolute position. Two predictions with identical attention patterns at different sequence positions produce similar features. We achieve this through three operations: max pooling over positions extracts peak attention magnitude without recording which position achieved it; entropy aggregates attention distributions into scalar sharpness measures; relative distances measure attention span rather than absolute indices.

\paragraph{Topology Preservation.}
Features preserve ordinal relationships needed for density-based clustering. The stability layer (earliest depth at which prediction converges) is encoded as a scalar rather than one-hot, placing predictions that converge at layer 1 closer to those at layer 2 than to those at layer 5 in feature space.

\subsection{Feature Architecture}
\label{sec:feature-arch}

Our architecture uses gated attention \citep{qiu2025gated}, where each head's output is modulated by a learned gate. The effective attention $\tilde{A}_h = A_h \cdot g_h$ reflects actual head contribution, with suppressed heads producing near-zero attention regardless of their softmax pattern. All features use effective attention, which correctly captures computational impact rather than nominal attention weights.

The feature system has two tiers. Tier 1 provides interpretable summary statistics for manual analysis. Tier 2 provides complete structural representation for clustering. Formal specifications appear in Appendix~\ref{appendix:feature-spec}.

\paragraph{Tier 1: Core Features (5D).}
\begin{center}
\small
\begin{tabular}{@{}lp{8cm}@{}}
\toprule
Feature & Definition \\
\midrule
\texttt{processing\_depth} & Layer $k^*$ at which prediction stabilizes \\
\texttt{confidence} & Final-layer probability $p^{(L-1)}(\hat{y})$ \\
\texttt{anchor\_mass} & Peak gated attention to syntactic heads \\
\texttt{entity\_mass} & Peak gated attention to entity heads \\
\texttt{context\_span} & Weighted mean attention distance \\
\bottomrule
\end{tabular}
\end{center}

\paragraph{Tier 2: Structural Features (163D).}
Complete representation organized into five components:

\begin{center}
\small
\begin{tabular}{@{}llp{6.5cm}@{}}
\toprule
Component & Dims & Content \\
\midrule
A. Trajectory & $3L{-}1$ & Layer-wise probability, margin over second-best, confidence drops \\
B. Stability & 2 & Convergence layer, max consecutive correct \\
C. Head Activation & $2LH$ & Peak gated attention at stable and final layers \\
D. Head Entropy & $2LH$ & Attention sharpness at stable and final layers \\
\bottomrule
\end{tabular}
\end{center}

For $L=6$ layers and $H=6$ heads, this yields 163 dimensions. Component C uses max pooling to achieve permutation invariance: $\max_{t_k} \tilde{A}_h[t_q, t_k]$ returns peak attention magnitude without recording which position achieved it. Component D uses entropy, which aggregates full attention distributions into scalar measures.

\subsection{Validation: The Raw Activation Baseline}
\label{sec:raw-baseline}

To confirm that engineered features capture computational structure rather than vocabulary, we cluster raw layer-5 activations (768 dimensions) and compute the Adjusted Rand Index \citep{hubert1985comparing} against our feature-based clustering. The ARI measures agreement between clusterings, with 0 indicating chance agreement and 1 indicating identity.

The result is ARI = 0.008. The clusterings share almost no structure. Raw activation clusters partition by token type: punctuation forms one cluster, function words another, verbs a third. Feature-based clusters partition by processing strategy: early versus late convergence, high versus low entity attention. The same token appears in different feature clusters depending on how the transformer processed its context.

\begin{figure}[t]
\centering
\includegraphics[width=0.9\columnwidth]{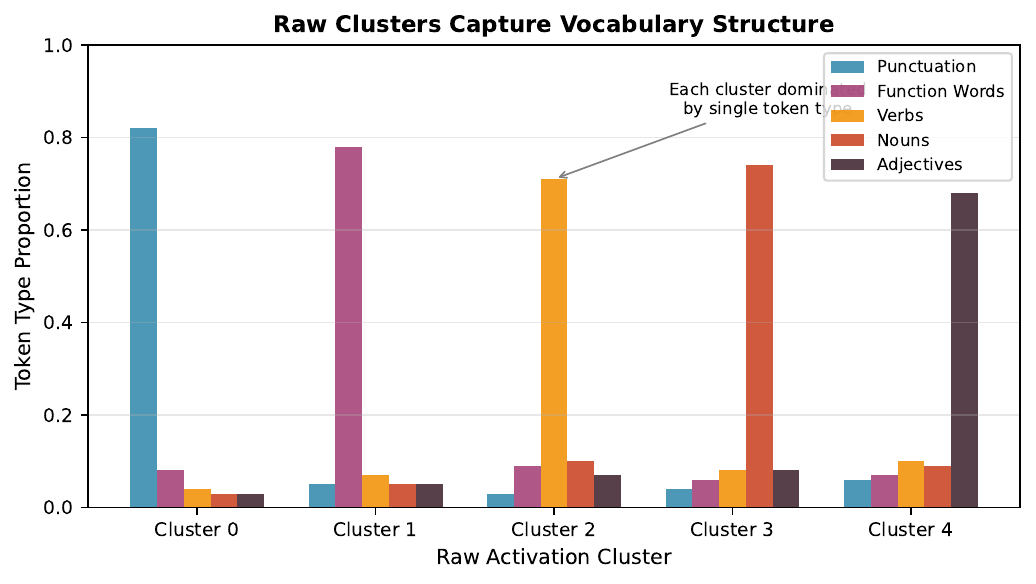}
\caption{\textbf{Raw activation clusters capture vocabulary, not computation.} Clustering layer-5 activations (768D) produces partitions dominated by token type. Our engineered features discover orthogonal structure validated through causal experiments (Section~\ref{sec:causal}).}
\label{fig:raw-clusters}
\end{figure}

Figure~\ref{fig:raw-clusters} visualizes this distinction. Raw clusters capture what tokens are; engineered features capture what the model does with them. This near-zero correlation validates that our features discover structure orthogonal to vocabulary, isolating computational dynamics from semantic content.

%% file: sections/03_architecture.tex
\section{The \cascade{} Architecture}
\label{sec:architecture}

Standard transformers permit gradient starvation: early-learned features dominate the loss landscape, causing later layers to develop redundant backup pathways rather than distinct functions \citep{pezeshki2021gradient}. The result is the Hydra effect. We introduce an architecture that constrains how layers can share computation, forcing specialization through three mechanisms: dual-stream processing that separates token identity from contextual accumulation, frozen symbolic streams that eliminate a primary channel for hidden redundancy, and per-layer supervision that provides independent gradient signal at each depth. Together with gated attention \citep{qiu2025gated}, these mechanisms create conditions where modularity can emerge.

The architecture serves as a positive control. If modular structure is achievable through training objectives alone, this architecture should reveal it. If the Hydra effect persists despite these constraints, modularity may be fundamentally incompatible with gradient-based training. Complete specifications appear in Appendix~\ref{appendix:architecture}.

\subsection{Dual-Stream Processing}
\label{sec:dual-stream}

The residual stream decomposes into two additive components:
\[
\mathbf{x} = \mathbf{x}_t + \mathbf{x}_e
\]

The token stream $\mathbf{x}_t$ initializes from token embeddings and maintains discrete symbolic identity. The contextual stream $\mathbf{x}_e$ starts at zero and accumulates context-dependent representations. This separation, inspired by DeBERTa's disentangled attention \citep{he2021deberta}, creates distinct pathways for token-level and context-level computation.

In standard dual-stream mode, attention updates $\mathbf{x}_t$ while feed-forward networks update $\mathbf{x}_e$. We employ a stricter variant.

\subsection{Frozen Symbolic Stream}
\label{sec:cascade}

In CASCADE mode, $\mathbf{x}_t$ remains frozen after initialization:
\[
\mathbf{x}_t^{(\ell)} = \mathbf{x}_t^{(0)} \quad \forall \ell
\]

Only $\mathbf{x}_e$ accumulates updates through the network. This eliminates the primary mechanism for hidden redundancy. Standard transformers can distribute computation across both streams, using the token stream as an additional channel for backup pathways. Freezing $\mathbf{x}_t$ forces the network to maintain separation between what tokens are (preserved in $\mathbf{x}_t$) and what context means (accumulated in $\mathbf{x}_e$).

The constraint has three consequences. Token identity remains intact through depth rather than being overwritten by contextual blending. Token-context integration delays to late layers, forcing early layers to operate on cleaner representations. Most critically, the effective capacity for distributed compensation reduces because the network cannot hide representations in a frozen stream.

\subsection{Per-Layer Supervision}
\label{sec:pls}

Per-layer supervision adds auxiliary prediction losses at each transformer layer. Standard transformers compute logits only at the final layer. We compute logits at every layer:
\[
\mathbf{z}^{(\ell)} = \text{lm\_head}(\text{LayerNorm}(\mathbf{h}^{(\ell)})) \quad \forall \ell \in \{0, \ldots, L-1\}
\]

The training objective combines final-layer loss with weighted auxiliary losses:
\[
\mathcal{L} = \mathcal{L}_{\text{CE}}(\mathbf{z}^{(L-1)}, y) + \lambda \sum_{\ell=0}^{L-2} w_\ell \cdot \mathcal{L}_{\text{CE}}(\mathbf{z}^{(\ell)}, y)
\]

We use linear decay weights $w_\ell = (\ell+1)/L$ to emphasize deeper layers while providing gradient signal throughout. The auxiliary weight $\lambda = 0.1$ balances supervision strength against final-layer optimization.

Per-layer supervision encourages each layer to make independent progress toward the prediction target. Without it, layers can defer to later corrections, creating redundant pathways where multiple layers perform similar functions. With it, each layer faces direct pressure to contribute distinctly. This provides training signal that exposes modularity if the architecture permits it, rather than forcing modularity through hard architectural constraints.

\subsection{Gated Attention}
\label{sec:gated-attention}

Following \citet{qiu2025gated}, each attention head includes a learned gate that modulates its output:
\[
\mathbf{y}_h = g_h \odot (\mathbf{A}_h \mathbf{V}_h)
\]

where $g_h \in [0,1]^{T \times d_h}$ is computed from the query representation via learned projection and sigmoid activation. The gate approaching zero effectively disables a head for specific query positions. The gate near one passes attention output unchanged.

Gated attention serves two functions. It regularizes attention toward discrete, interpretable patterns by encouraging heads to specialize through selective activation rather than contributing small amounts everywhere. Gate values also provide natural features for analysis: heads with consistently low gates can be identified as inactive for particular computational modes, reducing noise in our relational features.

\subsection{Control Model}
\label{sec:control}

To isolate per-layer supervision's contribution, we train a control model (C2) with identical architecture but without per-layer supervision. The control receives gradient signal only from the final layer, matching standard transformer training. Both models use CASCADE mode, dual streams, and gated attention. Both train on identical data with identical hyperparameters.

If per-layer supervision merely adds computational overhead without affecting internal structure, both models should show similar ablation sensitivity. If per-layer supervision exposes genuine modularity, the control should exhibit the Hydra effect while the supervised model reveals distinct causal pathways. Section~\ref{sec:causal} tests this prediction.

%% file: sections/04_validation.tex
\section{Validating Feature Sensitivity}
\label{sec:validation}

Our relational features should capture meaningful architectural differences if they isolate computational structure from vocabulary. We test this by comparing cluster statistics between the per-layer supervised model and the control. If features merely rediscover surface patterns, both architectures should produce similar clusterings. If features capture computational dynamics, the architectures should show distinct profiles.

\subsection{Experimental Setup}
\label{sec:cluster-setup}

We extract features from 10,000 token predictions across both models using identical text samples. For each prediction, we compute the full 163-dimensional Tier 2 feature vector: layer-wise probability trajectories, per-head peak attention, per-head entropy, and stability metrics. We cluster using $k$-means with $k=10$ on the full feature space and analyze the resulting depth distributions and functional profiles.

\subsection{Depth Distribution Comparison}
\label{sec:depth-comparison}

\begin{figure}[t]
\centering
\includegraphics[width=\columnwidth]{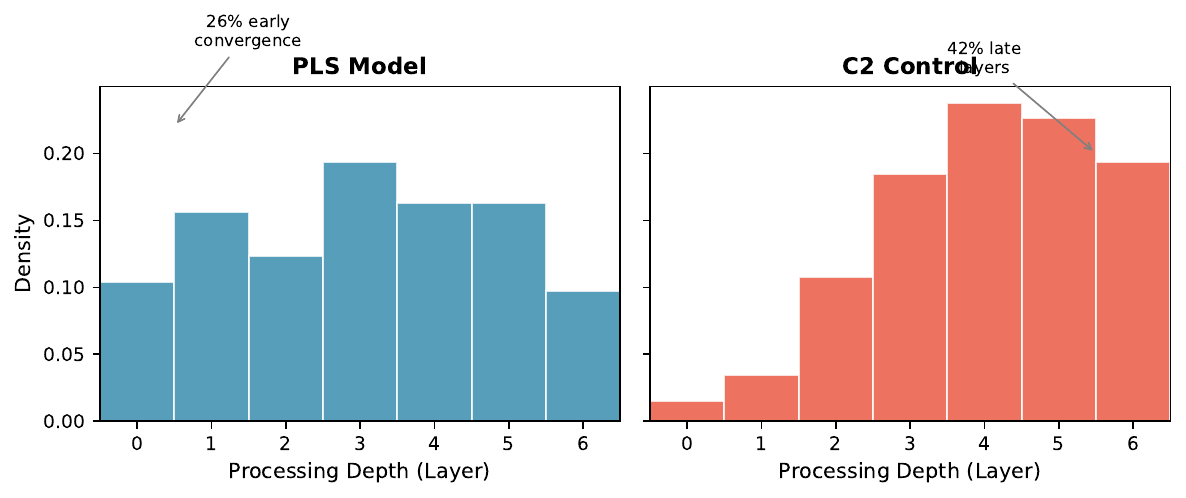}
\caption{\textbf{Depth distribution comparison.} PLS (left) shows bimodal structure: 26\% of predictions converge at layers 0--1, with a second peak at maximum depth. C2 (right) concentrates in middle-to-late layers.}
\label{fig:depth}
\end{figure}

Processing depth (the layer at which prediction stabilizes) provides clear architectural discrimination. Figure~\ref{fig:depth} shows the distributions; Table~\ref{tab:depth} summarizes the statistics.

\begin{table}[t]
\centering
\small
\begin{tabular}{@{}lrr@{}}
\toprule
\textbf{Depth Range} & \textbf{PLS} & \textbf{C2} \\
\midrule
Early (layers 0--1) & 26\% & 5\% \\
Middle (layers 2--4) & 48\% & 53\% \\
Late (layers 5--6) & 26\% & 42\% \\
\bottomrule
\end{tabular}
\caption{\textbf{Depth distribution by architecture.} PLS shows bimodal structure with substantial early convergence; C2 concentrates in middle-to-late layers.}
\label{tab:depth}
\end{table}

The most striking difference appears in early convergence. Per-layer supervision creates pressure for each layer to make progress, resulting in 26\% of predictions converging at layers 0--1 compared to only 5\% for the control. This bimodal pattern distinguishes predictions that require minimal context (converging immediately) from those requiring full processing depth.

The control model concentrates computation in later layers (42\% at layers 5--6 versus 26\% for PLS). Without per-layer supervision, the network defers decisions and distributes computation redundantly across layers. This deferred computation pattern enables the Hydra effect by maintaining multiple pathways to the same prediction.

\subsection{Cluster Functional Profiles}
\label{sec:cluster-profiles}

Beyond depth statistics, we characterize cluster contents through systematic sample inspection. For each cluster, we extract 15 diverse samples maximizing within-cluster distance and analyze their linguistic properties.

Per-layer supervision produces clusters with distinct computational profiles. Cluster 2 contains section headers and tutorial introductions with high anchor attention and early convergence. Cluster 5 contains pronoun antecedents and coreference cases with high entity attention and late convergence. Cluster 7 contains conjunctions and relative pronouns with distributed attention and middle-layer convergence. Cluster 9 contains ambiguous continuations with low confidence and maximum depth.

The control model produces more uniform profiles. Depth distributions are similar across clusters, and attention patterns show less differentiation. The same clustering procedure applied to both architectures produces sharper functional boundaries in the supervised model.

\subsection{Raw Activation Baseline}
\label{sec:raw-validation}

As reported in Section~\ref{sec:raw-baseline}, clustering raw layer-5 activations produces structure orthogonal to our engineered features with adjusted Rand index 0.008. Raw clusters partition by token identity (punctuation, verbs, nouns). Our clusters partition by processing strategy (early versus late convergence, high versus low entity attention). The same word appears in different clusters depending on how the transformer processed its context.

This validates that relational features discover computational structure invisible to content-based representations, meeting the requirement for distinguishing computational modes from semantic categories.

%% file: sections/05_causal.tex
\section{Causal Validation}
\label{sec:causal}

Observational analysis cannot distinguish accessible information from used information \citep{belinkov2022probing}. We validate our claims through targeted ablation: if per-layer supervision creates genuine modularity, ablating identified heads should produce measurable behavioral changes. If the Hydra effect persists, ablation damage should be absorbed through distributed compensation.

\subsection{Experimental Design}
\label{sec:ablation-design}

We follow the causal abstraction framework \citep{geiger2021causal}: a model component is causally responsible for a behavior if intervening on that component changes the behavior. We ablate specific attention heads by zeroing their output and measure the effect on task performance.

We evaluate on three tasks requiring different computational strategies. Capitalization preservation requires predicting the correct surface form given context establishing capitalization (8 cases). Gender resolution requires resolving pronouns in contexts with gendered antecedents (16 cases). Winograd schema cases require disambiguating pronouns through world knowledge \citep{levesque2012winograd} (50 cases). Based on attention pattern analysis, we identify three functionally distinct head groups in layer 5: entity heads tracking entity mentions, anchor heads attending to syntactic anchors, and semantic heads integrating content.

For each test case, we compute the change in probability assigned to the correct answer:
\[
\Delta = P_{\text{ablated}}(\text{correct}) - P_{\text{baseline}}(\text{correct})
\]
Negative values indicate the ablated heads contributed to correct prediction. Positive values indicate they interfered.

\subsection{Results}
\label{sec:ablation-results}

Table~\ref{tab:ablation} presents ablation effects across tasks and models. Per-layer supervision produces effects 5 to 23 times larger than controls.

\begin{table}[t]
\centering
\small
\begin{tabular}{@{}llrrr@{}}
\toprule
\textbf{Task} & \textbf{Heads} & \textbf{PLS} $\Delta$ & \textbf{C2} $\Delta$ & \textbf{Ratio} \\
\midrule
Capitalization & Entity (h54+h55) & $-$11.4\% & $-$2.1\% & 5.4$\times$ \\
Gender & Anchor (h52+h53) & $-$2.0\% & $-$0.3\% & 6.7$\times$ \\
Winograd & Entity (h54+h55) & $+$1.15\% & $+$0.05\% & 23$\times$ \\
\bottomrule
\end{tabular}
\caption{\textbf{Ablation effects by task and model.} PLS heads produce 5--23$\times$ larger effects than C2 heads. Negative values indicate contribution; positive values indicate interference.}
\label{tab:ablation}
\end{table}

We focus detailed analysis on capitalization, where the effect is both largest and most mechanistically transparent. Ablating entity heads in the supervised model reduces capitalization accuracy by 11.4 percent. The model loses its ability to copy surface form from context. The same ablation in the control produces only 2.1 percent degradation.

\begin{figure}[t]
\centering
\includegraphics[width=0.9\columnwidth]{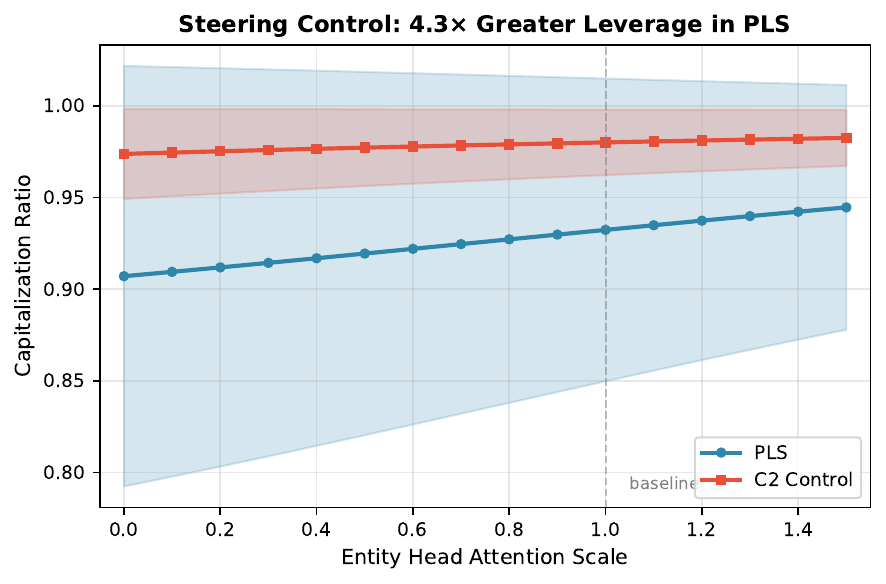}
\caption{\textbf{Capitalization steering via attention scaling.} PLS (blue) shows 4$\times$ greater control range than C2 (orange). Scaling entity head attention from 0 to 1.5$\times$ produces smooth, monotonic changes in capitalization probability.}
\label{fig:steering}
\end{figure}

Figure~\ref{fig:steering} shows the full control range. Scaling entity head attention from zero to 1.5 times baseline produces smooth, monotonic changes in capitalization probability. Amplifying entity heads increases capitalization preservation; suppressing them decreases it. This provides four times greater control leverage than the control model, enabling surgical steering of behavior.

Extended validation on 50 Winograd schema cases reveals the signature of exposed versus hidden circuitry. Per-layer supervision produces ablation effect mean 1.15 percent with standard deviation 6.32 percent. The control produces mean 0.05 percent with standard deviation 0.63 percent. Figure~\ref{fig:winograd-dist} visualizes this contrast. The control shows a tight cluster around zero. The Hydra actively compensates for ablation damage, suppressing variance. The supervised model shows wide spread. The system reveals which cases depend on the ablated circuits.

The positive mean indicates entity heads slightly interfere with Winograd performance. They may be applying entity-tracking heuristics that conflict with deeper reasoning requirements. This interference is visible in the supervised model and invisible in the control. The 23 times ratio in effect magnitude, combined with 10 times ratio in variance, provides the signature of unmasked modularity.

\subsection{Functional Reorganization}
\label{sec:reorganization}

\begin{figure}[t]
\centering
\begin{subfigure}[t]{\columnwidth}
\centering
\includegraphics[width=\columnwidth]{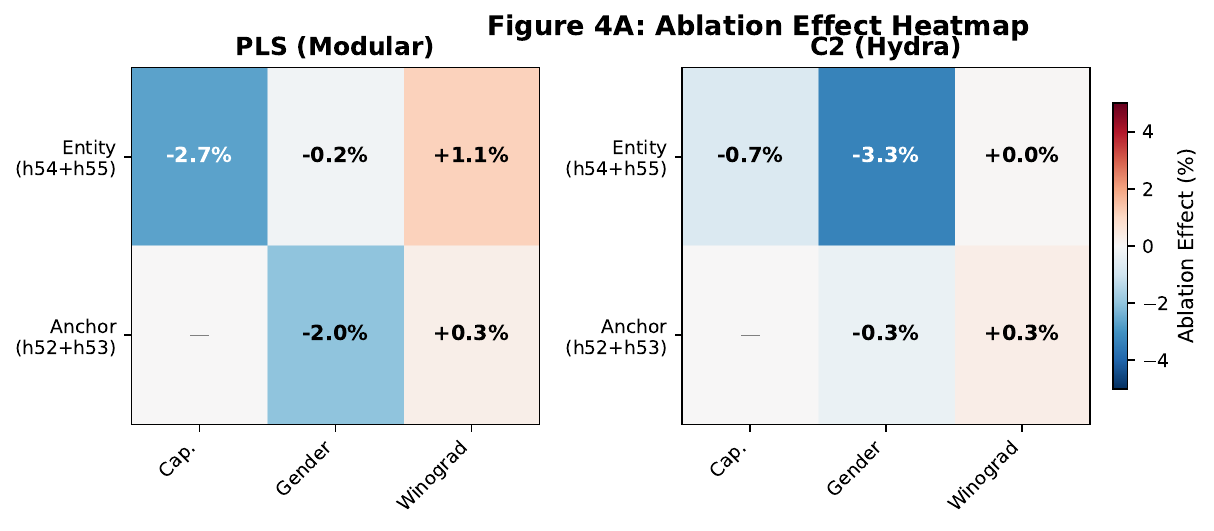}
\caption{\textbf{Ablation effect matrix.} PLS (left) shows diagonal sparsity: entity heads drive capitalization, anchor heads drive gender. C2 (right) shows functional smearing.}
\label{fig:heatmap}
\end{subfigure}
\vspace{0.5em}
\begin{subfigure}[t]{\columnwidth}
\centering
\includegraphics[width=0.85\columnwidth]{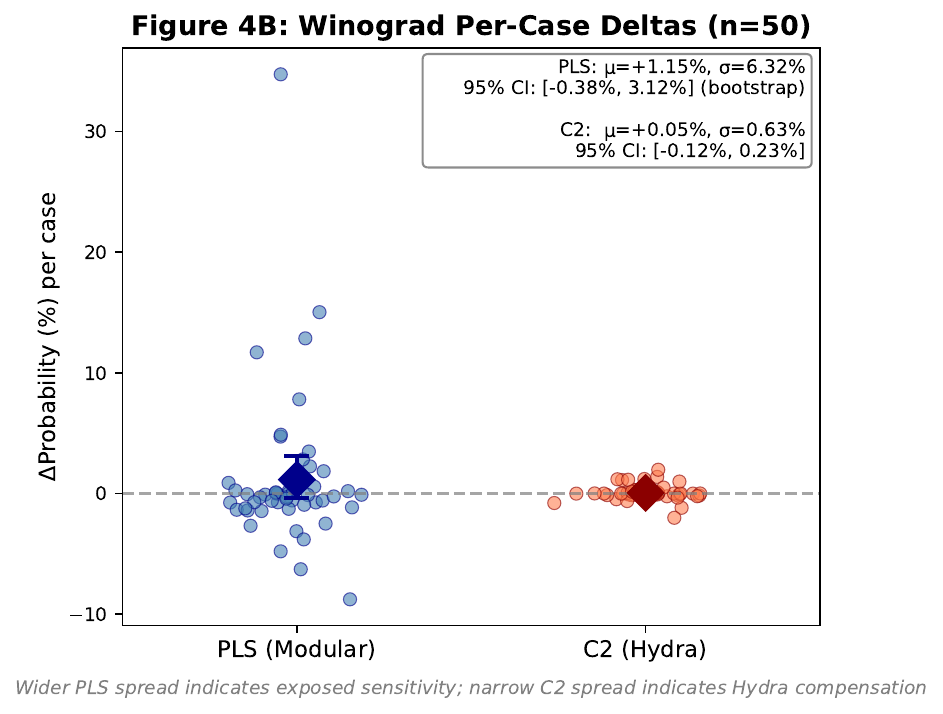}
\caption{\textbf{Winograd per-case effects (n=50).} C2 clusters tightly around zero ($\sigma=0.63\%$). PLS spreads widely ($\sigma=6.32\%$). Wider variance reveals exposed circuitry.}
\label{fig:winograd-dist}
\end{subfigure}
\caption{\textbf{The modularity signature.} Task-specific routing in PLS versus entangled routing in C2. Variance difference reveals exposed versus hidden circuitry.}
\label{fig:winograd}
\end{figure}

Figure~\ref{fig:heatmap} reveals architectural distinction through task-specific routing. In the supervised model, entity heads drive capitalization with minimal effect on gender. Anchor heads drive gender with minimal effect on capitalization. This diagonal sparsity indicates functional specialization where different tasks route through different heads.

In the control, the pattern smears. Entity heads impact gender more than capitalization. This functional entanglement is the substrate of the Hydra effect. When computation distributes across redundant pathways, ablating any pathway causes unpredictable collateral damage.

\subsection{Interpretation}
\label{sec:causal-interpretation}

These results validate that per-layer supervision converts silent interference into exposed sensitivity. The 5 to 23 times ablation effect ratios demonstrate three claims. First, modularity exists: specific heads perform specific functions. Second, modularity is verifiable: ablation produces predictable, task-specific effects rather than diffuse degradation. Third, modularity enables control: the four times leverage on capitalization shows we can surgically steer behavior without destroying other capabilities.

The features from Section~\ref{sec:features} predict which tokens will show high ablation sensitivity, closing the loop between observational analysis and causal validation. The Hydra effect is not inevitable. With appropriate architectural support, transformer computation organizes into interpretable, controllable modules.

%% file: sections/06_discussion.tex
\section{Discussion}
\label{sec:discussion}

\subsection{Implications for Interpretability Research}
\label{sec:implications}

Our feature methodology and architectural interventions are mutually reinforcing. Relational features reveal structure that per-layer supervision creates. Per-layer supervision provides controlled validation for the feature methodology. This mutual reinforcement distinguishes our approach from post-hoc interpretability methods where features might simply rediscover patterns the researcher expects to find. When features predict causal effects on an independently designed architecture, the features capture genuine computational structure rather than measurement artifacts.

Three methodological implications follow. First, the raw activation baseline with adjusted Rand index 0.008 should become standard practice. Any activation clustering approach should demonstrate that it discovers structure orthogonal to vocabulary rather than rediscovering the embedding space. Our three invariances provide a template for principled feature design that isolates computational dynamics from semantic content.

Second, interpretability can be engineered rather than merely discovered. Rather than accepting the Hydra effect as inevitable, we show that training objectives can expose modular structure. Per-layer supervision does not force modularity through hard architectural constraints but creates conditions where specialization emerges naturally. This suggests a research program designing architectures where modularity is a training objective rather than an emergent property to be excavated retrospectively.

Third, mode discovery enables targeted intervention. The 5 to 23 times ablation effect ratios demonstrate that identified modes have predictable behavioral consequences. This transforms interpretability from passive observation to active control. Rather than noting that a head attends to entities, we can scale that head by 1.5 times and increase capitalization preservation by measurable amounts. The smooth steering curves indicate that behavior can be titrated rather than merely toggled.

\subsection{Safety Implications}
\label{sec:safety}

Surgical control without retraining has direct safety applications. If a head implements an undesired behavior, it can be ablated or downweighted at inference time. The smooth steering curves suggest that behavior can be adjusted gradually rather than eliminated entirely, permitting fine-grained control over model outputs.

The interference finding on Winograd cases reveals that entity heads slightly degrade reasoning performance, likely by applying surface-level heuristics that conflict with deeper requirements. This interference is invisible in standard models due to compensation but becomes detectable under per-layer supervision. Identifying which components interfere with which tasks provides actionable information for model improvement.

However, the interference finding also indicates caution. Heads optimized for one task may degrade performance on others. Surgical interventions require understanding task scope and potential collateral effects. The functional reorganization results show that per-layer supervision creates cleaner task-specific routing, but the control model's entangled routing may provide robustness through redundancy. Trading interpretability for redundancy represents a fundamental tradeoff requiring careful evaluation in deployment contexts.

\subsection{Limitations}
\label{sec:limitations}

We acknowledge four limitations. First, results are demonstrated on CASCADE and per-layer supervision variants. Whether the modularity signatures persist in standard transformer architectures or other architectural families requires further study. The control model uses CASCADE mode, which already constrains computation relative to fully standard transformers.

Second, our models have 6 layers and 6 attention heads per layer. Whether modularity signatures persist at larger scales is unresolved. Larger models may develop more sophisticated compensation mechanisms that resist per-layer supervision, or the supervision signal may scale favorably as model capacity increases.

Third, we train on grade-school instructional text rather than web-scale corpora. This controlled domain permits clean isolation of computational strategies but may not reflect the full complexity of natural language distribution. We demonstrate foundational methodology in a setting where causal signals can be clearly measured before scaling to environments with multiple confounds.

Fourth, we do not directly compare to sparse autoencoder methods \citep{bricken2023monosemanticity}. Our features capture computational dynamics rather than semantic features, making direct comparison non-trivial. Sparse autoencoders decompose what representations encode, while our features track how computation unfolds. The approaches are complementary rather than competing, but systematic comparison would strengthen both methodologies.

%% file: sections/07_related.tex
\section{Related Work}
\label{sec:related}

\paragraph{The Hydra effect and attention redundancy.}
\citet{mcgrath2023hydra} formalized the Hydra effect, demonstrating that ablated attention heads are compensated by downstream MLPs through emergent self-repair. Earlier work established the structural basis: \citet{michel2019sixteen} showed that most attention heads can be pruned at test time with minimal degradation, while \citet{voita2019analyzing} found that only specialized heads perform critical functions. Our work addresses this redundancy directly through per-layer supervision that exposes hidden compensation mechanisms rather than accepting them as inevitable.

\paragraph{Mechanistic interpretability.}
The circuits framework \citep{elhage2021mathematical} established the residual stream as a shared communication channel enabling distributed computation. \citet{olsson2022context} identified induction heads as distributed mechanisms for in-context learning. \citet{wang2023interpretability} traced the indirect object identification circuit in GPT-2, discovering backup heads that activate when primary heads are ablated. \citet{bricken2023monosemanticity} introduced sparse autoencoders to decompose polysemantic neurons into interpretable features. Our relational features are complementary: sparse autoencoders reveal what representations encode, while our features capture how computation unfolds across layers. The raw activation baseline validates that our features discover structure orthogonal to semantic content.

\paragraph{Attention analysis and causal methods.}
\citet{clark2019bert} identified attention heads tracking syntax and coreference, but \citet{jain2019attention} showed that attention weights correlate weakly with feature importance. \citet{wiegreffe2019attention} clarified conditions under which attention can be explanatory. Our causal validation through ablation and steering experiments demonstrates that identified heads have predictable behavioral consequences rather than merely correlating with outputs. \citet{geiger2021causal} defined causal abstraction as the condition where neural components implement interpretable causal models. \citet{geiger2022inducing} proposed Interchange Intervention Training to induce causal structure. Our approach differs by using per-layer supervision to create general conditions for modularity rather than training for specific causal structures.

\paragraph{Architectural approaches to interpretability.}
Mixture of experts \citep{shazeer2017outrageously} achieves modularity through explicit routing, but the router itself becomes opaque. \citet{liu2024bimt} proposed brain-inspired modular training with locality constraints. \citet{he2021deberta} introduced disentangled attention separating position and content. Our CASCADE mode extends this separation to maintain frozen symbolic streams throughout depth. Per-layer supervision achieves modularity within standard transformer blocks without explicit routing mechanisms, using training objectives alone to expose functional specialization.

\paragraph{Gradient dynamics and training.}
\citet{pezeshki2021gradient} identified gradient starvation: once dominant features minimize loss, gradients for alternative features vanish. This explains why standard training produces redundant representations. Per-layer supervision counteracts starvation by providing independent gradient signal at each depth, preventing the accumulation of backup circuits that enables the Hydra effect.

%% file: sections/08_conclusion.tex
\section{Conclusion}
\label{sec:conclusion}

We demonstrate that architectural interventions can expose computational modularity that standard training obscures. Per-layer supervision converts the Hydra effect from silent compensation into measurable sensitivity, producing ablation effects 5 to 23 times larger than architecturally identical controls. This enables surgical control: scaling identified attention heads produces smooth, predictable changes in model behavior with 4 times greater leverage than control models.

Three contributions establish this methodology. First, we engineer features that capture computational dynamics rather than vocabulary structure, validated by near-zero correlation with raw activation clustering. These features distinguish how the model processes tokens from what tokens appear in context. Second, we demonstrate that per-layer supervision forces functional reorganization where different tasks route through different attention heads, contrasting with the entangled routing that enables distributed compensation in control models. Third, we provide causal validation through targeted ablation and steering experiments that predict behavioral changes from feature analysis.

The approach establishes that modularity can be engineered through training objectives rather than merely discovered through post-hoc analysis. Standard training produces redundant pathways because gradient-based optimization encounters no pressure toward interpretability. Per-layer supervision creates conditions where specialization emerges by providing independent gradient signal at each depth.

We do not claim this methodology scales immediately to billion-parameter models or generalizes beyond the architectural family we study. Our models have 6 layers trained on controlled text rather than web-scale corpora. Whether per-layer supervision exposes modularity in larger models operating on natural language distribution requires systematic investigation. We demonstrate foundational methodology for verifying that identified computational modes have causal consequences, establishing requirements for transforming interpretability from passive observation to active control.

%% file: sections/appendix_architecture.tex

\section{\cascade{} Architecture Specification}
\label{appendix:architecture}

This appendix provides complete mathematical specifications for the \cascade{} architecture described in Section~\ref{sec:architecture}.

\subsection{Notation}

\begin{center}
\begin{tabular}{@{}lll@{}}
\toprule
Symbol & Description & Value \\
\midrule
$L$ & Number of transformer layers & 6 \\
$H$ & Attention heads per layer & 6 \\
$d$ & Hidden dimension & 384 \\
$d_h$ & Head dimension ($d/H$) & 64 \\
$d_{\text{ff}}$ & Feed-forward intermediate dimension & 1536 \\
$V$ & Vocabulary size (GPT-2 tokenizer) & 50,257 \\
$T$ & Sequence length & Variable \\
\bottomrule
\end{tabular}
\end{center}

\subsection{Dual-Stream Decomposition}

\subsubsection{Stream Definitions}

The residual stream at layer $\ell$ is decomposed:
\[
\mathbf{x}^{(\ell)} = \mathbf{x}_t^{(\ell)} + \mathbf{x}_e^{(\ell)}
\]

\paragraph{Token stream initialization.}
\[
\mathbf{x}_t^{(0)} = \mathbf{W}_E[t] \in \mathbb{R}^d
\]
where $\mathbf{W}_E \in \mathbb{R}^{V \times d}$ is the token embedding matrix and $t$ is the token index.

\paragraph{Contextual stream initialization.}
\[
\mathbf{x}_e^{(0)} = \mathbf{0} \in \mathbb{R}^d
\]

\subsubsection{Stream Update Rules}

In standard dual-stream mode:
\begin{align}
\mathbf{x}_t^{(\ell+1)} &= \mathbf{x}_t^{(\ell)} + \text{Attn}^{(\ell)}(\mathbf{x}^{(\ell)}) \\
\mathbf{x}_e^{(\ell+1)} &= \mathbf{x}_e^{(\ell)} + \text{FFN}^{(\ell)}(\mathbf{x}^{(\ell)})
\end{align}

Attention updates the token stream; feed-forward networks update the contextual stream.

\subsection{CASCADE Mode: Frozen Symbolic Stream}

In CASCADE mode, the token stream remains frozen after initialization:
\begin{align}
\mathbf{x}_t^{(\ell)} &= \mathbf{x}_t^{(0)} \quad \forall \ell \in \{0, \ldots, L-1\} \\
\mathbf{x}_e^{(\ell+1)} &= \mathbf{x}_e^{(\ell)} + \text{Attn}^{(\ell)}(\mathbf{x}^{(\ell)}) + \text{FFN}^{(\ell)}(\mathbf{x}^{(\ell)})
\end{align}

All updates accumulate in the contextual stream. The combined representation at layer $\ell$ is:
\[
\mathbf{x}^{(\ell)} = \mathbf{x}_t^{(0)} + \mathbf{x}_e^{(\ell)}
\]

\paragraph{Rationale.}
The frozen token stream preserves discrete symbolic identity through depth. In standard transformers, token embeddings are progressively overwritten by contextual information, enabling distributed compensation (the Hydra effect). CASCADE mode forces the network to maintain separation between ``what the token is'' ($\mathbf{x}_t$) and ``what the context means'' ($\mathbf{x}_e$).

\subsection{Gated Attention}

\subsubsection{Standard Scaled Dot-Product Attention}

For head $h$ at layer $\ell$:
\begin{align}
\mathbf{Q}_h &= \mathbf{x}^{(\ell)} \mathbf{W}^Q_h, \quad \mathbf{K}_h = \mathbf{x}^{(\ell)} \mathbf{W}^K_h, \quad \mathbf{V}_h = \mathbf{x}^{(\ell)} \mathbf{W}^V_h \\
\mathbf{A}_h &= \text{softmax}\left(\frac{\mathbf{Q}_h \mathbf{K}_h^\top}{\sqrt{d_h}} + \mathbf{M}\right)
\end{align}

where $\mathbf{M}$ is the causal mask ($-\infty$ for future positions, $0$ otherwise).

\subsubsection{Output Gating}

Following \citet{qiu2025gated}, we add a learned gate that modulates each head's output:
\[
\mathbf{y}_h = g_h \odot (\mathbf{A}_h \mathbf{V}_h)
\]

The gate $g_h \in [0,1]^{T \times d_h}$ is computed:
\[
g_h = \sigma(\mathbf{Q}_h \mathbf{W}^G_h + \mathbf{b}^G_h)
\]

where $\mathbf{W}^G_h \in \mathbb{R}^{d_h \times d_h}$ and $\mathbf{b}^G_h \in \mathbb{R}^{d_h}$ are learned parameters, and $\sigma$ is the sigmoid function.

\paragraph{Effective (gated) attention.}
For analysis, we define the effective attention weights:
\[
\tilde{A}_h[t_q, t_k] = A_h[t_q, t_k] \cdot \bar{g}_h[t_q]
\]

where $\bar{g}_h[t_q] = \frac{1}{d_h}\sum_i g_h[t_q, i]$ is the mean gate value across the head dimension. This captures the effective contribution of each attention pattern to the output.

\subsubsection{Multi-Head Aggregation}

Head outputs are concatenated and projected:
\[
\text{Attn}^{(\ell)}(\mathbf{x}) = \text{Concat}(\mathbf{y}_1, \ldots, \mathbf{y}_H) \mathbf{W}^O
\]

where $\mathbf{W}^O \in \mathbb{R}^{d \times d}$ is the output projection.

\subsection{Per-Layer Supervision}

\subsubsection{Per-Layer Logits}

At each layer $\ell$, we compute logits:
\[
\mathbf{z}^{(\ell)} = \text{lm\_head}(\text{LayerNorm}(\mathbf{x}^{(\ell)}))
\]

The language modeling head is shared across all layers:
\[
\text{lm\_head}(\mathbf{h}) = \mathbf{h} \mathbf{W}_E^\top
\]

using tied embeddings ($\mathbf{W}_E$ is the token embedding matrix).

\subsubsection{Training Objective}

The loss combines final-layer cross-entropy with weighted auxiliary losses:
\[
\mathcal{L} = \mathcal{L}_{\text{CE}}(\mathbf{z}^{(L-1)}, y) + \lambda \sum_{\ell=0}^{L-2} w_\ell \cdot \mathcal{L}_{\text{CE}}(\mathbf{z}^{(\ell)}, y)
\]

\paragraph{Weight schedule.}
Linear decay weights emphasize deeper layers:
\[
w_\ell = \frac{\ell + 1}{L}
\]

For $L=6$: $w_0 = \frac{1}{6}$, $w_1 = \frac{2}{6}$, $w_2 = \frac{3}{6}$, $w_3 = \frac{4}{6}$, $w_4 = \frac{5}{6}$.

\paragraph{Auxiliary weight.}
We use $\lambda = 0.1$. This balances providing gradient signal to early layers against over-constraining the final prediction.

\subsection{Feed-Forward Networks}

Standard two-layer MLP with GELU activation:
\[
\text{FFN}(\mathbf{x}) = \text{GELU}(\mathbf{x} \mathbf{W}_1 + \mathbf{b}_1) \mathbf{W}_2 + \mathbf{b}_2
\]

where $\mathbf{W}_1 \in \mathbb{R}^{d \times d_{\text{ff}}}$, $\mathbf{W}_2 \in \mathbb{R}^{d_{\text{ff}} \times d}$.

\subsection{Layer Normalization}

We use Pre-LN (layer normalization before attention and FFN):
\begin{align}
\mathbf{a}^{(\ell)} &= \text{Attn}^{(\ell)}(\text{LN}(\mathbf{x}^{(\ell)})) \\
\mathbf{f}^{(\ell)} &= \text{FFN}^{(\ell)}(\text{LN}(\mathbf{x}^{(\ell)} + \mathbf{a}^{(\ell)}))
\end{align}

\subsection{Training Configuration}

\begin{center}
\begin{tabular}{@{}ll@{}}
\toprule
Hyperparameter & Value \\
\midrule
Optimizer & AdamW \\
Learning rate & $3 \times 10^{-4}$ \\
Weight decay & 0.1 \\
Batch size & 64 \\
Sequence length & 512 \\
Warmup steps & 1000 \\
LR schedule & Cosine decay \\
Gradient clipping & 1.0 \\
Dropout & 0.1 \\
\bottomrule
\end{tabular}
\end{center}

\subsection{Model Variants}

\begin{center}
\begin{tabular}{@{}lcccc@{}}
\toprule
Model & Dual-Stream & CASCADE & PLS & Gated Attn \\
\midrule
\cascade{} (full) & \checkmark & \checkmark & \checkmark & \checkmark \\
C2 (control) & \checkmark & \checkmark & --- & \checkmark \\
\bottomrule
\end{tabular}
\end{center}

The control model C2 differs only in lacking per-layer supervision, isolating PLS's contribution to modularity.

%% file: sections/appendix_feature_specification.tex

\section{Complete Feature Specification}
\label{appendix:feature-spec}

This appendix provides the complete mathematical specification of the feature extraction system used throughout this paper. We define notation, specify each feature component, and document implementation details.

\subsection{Notation and Conventions}
\label{appendix:notation}

\subsubsection{Architectural Parameters}

\begin{center}
\begin{tabular}{@{}lll@{}}
\toprule
Symbol & Description & Values in this paper \\
\midrule
$L$ & Number of layers & 6 \\
$H$ & Heads per layer & 6 \\
$T$ & Sequence length (context) & Variable \\
$V$ & Vocabulary size & 50,257 (GPT-2 tokenizer) \\
$d$ & Hidden dimension & 384 \\
$d_h$ & Head dimension ($d/H$) & 64 \\
\bottomrule
\end{tabular}
\end{center}

\textbf{Note:} We use a custom transformer architecture with GPT-2's BPE tokenizer. The architecture parameters ($L=6$, $H=6$, $d=384$) are smaller than GPT-2 to enable controlled experimentation; the tokenizer provides standard subword segmentation.

\subsubsection{Index Conventions}

\begin{itemize}[nosep]
\item Layer indices: $\ell \in \{0, 1, \ldots, L-1\}$
\item Head indices: $h \in \{0, 1, \ldots, H-1\}$
\item Position indices: $t \in \{0, 1, \ldots, T-1\}$
\item Query position for predicting token $t$: $t_q = t - 1$ (previous token position)
\end{itemize}

\subsubsection{Tensor Shapes}

\begin{center}
\begin{tabular}{@{}llp{6cm}@{}}
\toprule
Tensor & Shape & Description \\
\midrule
$\mathbf{Z}^{(\ell)}$ & $[T, V]$ & Logits at layer $\ell$ \\
$\mathbf{A}^{(\ell)}$ & $[H, T, T]$ & Standard attention weights at layer $\ell$ \\
$\tilde{\mathbf{A}}^{(\ell)}$ & $[H, T, T]$ & Effective (gated) attention at layer $\ell$ \\
$g^{(\ell)}_h$ & $[T]$ & Gate values for head $h$ at layer $\ell$ \\
\bottomrule
\end{tabular}
\end{center}

\subsection{Gated Attention}
\label{appendix:gated-attention}

Our architecture uses gated attention \citep{qiu2025gated}, where each head's output is modulated by a learned gate:

\paragraph{Standard attention.}
\[
A^{(\ell)}_{h}[t_q, t_k] = \text{softmax}\left(\frac{\mathbf{q}_{t_q}^\top \mathbf{k}_{t_k}}{\sqrt{d_h}}\right), \quad \sum_{t_k \leq t_q} A^{(\ell)}_{h}[t_q, t_k] = 1
\]

\paragraph{Gated (effective) attention.}
\[
\tilde{A}^{(\ell)}_{h}[t_q, t_k] = A^{(\ell)}_{h}[t_q, t_k] \cdot g^{(\ell)}_{h}[t_q], \quad \sum_{t_k \leq t_q} \tilde{A}^{(\ell)}_{h}[t_q, t_k] = g^{(\ell)}_{h}[t_q] \leq 1
\]

The gate $g^{(\ell)}_{h}[t_q] \in [0,1]$ is computed via sigmoid from a learned projection of the query representation. All attention-based features in this paper use \textbf{effective (gated) attention} $\tilde{A}$, which correctly reflects the actual contribution of each head to the computation.

\paragraph{Notation convention.} In annotation outputs (e.g., \texttt{(a52->0.87)}), values are gated attention weights, not standard softmax probabilities. A value of 0.08 may indicate either weak attention or a suppressed gate.

\subsection{Design Principles}
\label{appendix:design-principles}

Three invariances govern feature design:

\paragraph{Token Identity Invariance.}
Features must not depend on which tokens appear. Two predictions with identical computational signatures cluster together whether the context contains ``cat'' or ``dog.''

\paragraph{Permutation Invariance.}
Features must not depend on absolute position. Two predictions with identical computational patterns at different sequence positions produce similar feature vectors. This is achieved by:
\begin{itemize}[nosep]
\item Using $\max$ over positions (discards which position achieved max)
\item Using entropy (aggregates over all positions)
\item Using relative distances (attention span), not absolute indices
\end{itemize}

\paragraph{Topology Preservation.}
Features preserve ordinal relationships for manifold learning. Stability layer $k^*$ is encoded as a scalar, not one-hot, so that layer 1 is ``closer'' to layer 2 than to layer 5 in feature space.

\subsection{Feature Tiers}
\label{appendix:tiers}

\subsubsection{Tier 1: Core Features (5 dimensions)}

Minimal interpretable features summarizing the computational signature:

\begin{center}
\begin{tabular}{@{}lcp{7cm}@{}}
\toprule
Feature & Dims & Definition \\
\midrule
\texttt{processing\_depth} & 1 & Number of layers until prediction stabilizes \\
\texttt{confidence\_score} & 1 & $p^{(L-1)}_t = \text{softmax}(\mathbf{Z}^{(L-1)}[t])[\hat{y}]$ \\
\texttt{anchor\_mass} & 1 & $\sum_{h \in \mathcal{H}_A} \max_{t_k < t_q} \tilde{A}^{(\ell_A)}_{h}[t_q, t_k]$ \\
\texttt{entity\_mass} & 1 & $\sum_{h \in \mathcal{H}_E} \max_{t_k < t_q} \tilde{A}^{(\ell_E)}_{h}[t_q, t_k]$ \\
\texttt{context\_span} & 1 & Weighted mean attention distance at final layer \\
\bottomrule
\end{tabular}
\end{center}

Here $\mathcal{H}_A = \{(5,2), (5,3)\}$ are anchor heads and $\mathcal{H}_E = \{(5,4), (5,5)\}$ are entity heads, identified through attention pattern analysis.

\subsubsection{Tier 2: Structural Features (163 dimensions)}

Complete structural representation for clustering:

\begin{center}
\begin{tabular}{@{}lcp{7cm}@{}}
\toprule
Component & Dims & Description \\
\midrule
\textbf{A. Trajectory} & $3L-1$ & \\
\quad \texttt{traj\_prob} & $L$ & $p^{(\ell)}_t$ if prediction correct, else 0 \\
\quad \texttt{traj\_margin} & $L$ & $p^{(\ell)}_t - p^{(\ell)}_{t,2}$ (margin over second-best) \\
\quad \texttt{traj\_drops} & $L-1$ & $\max(0, p^{(\ell-1)}_t - p^{(\ell)}_t)$ \\
\midrule
\textbf{B. Stability} & 2 & \\
\quad $k^*$ & 1 & Stability layer (scalar, ordinal) \\
\quad $\kappa$ & 1 & Max consecutive correct layers \\
\midrule
\textbf{C. Head Activation} & $2 \cdot L \cdot H$ & \\
\quad \texttt{head\_act\_stable} & $L \cdot H$ & $\max_{t_k < t_q} \tilde{A}^{(\ell)}_{h}[t_q, t_k]$ at $\ell = k^*$ \\
\quad \texttt{head\_act\_final} & $L \cdot H$ & Same at $\ell = L-1$ \\
\midrule
\textbf{D. Head Entropy} & $2 \cdot L \cdot H$ & \\
\quad \texttt{head\_ent\_stable} & $L \cdot H$ & $H(\tilde{A}^{(\ell)}_{h} / \sum \tilde{A})$ at $\ell = k^*$ \\
\quad \texttt{head\_ent\_final} & $L \cdot H$ & Same at $\ell = L-1$ \\
\bottomrule
\end{tabular}
\end{center}

\textbf{Total:} $(3 \cdot 6 - 1) + 2 + (2 \cdot 6 \cdot 6) + (2 \cdot 6 \cdot 6) = 17 + 2 + 72 + 72 = 163$ dimensions.

\subsubsection{Optional: Position Features (+4 dimensions)}

Component E adds attention geometry features that partially break permutation invariance to capture local vs.\ global attention patterns:

\begin{center}
\begin{tabular}{@{}lcp{7cm}@{}}
\toprule
Feature & Dims & Definition \\
\midrule
\texttt{attn\_span\_stable} & 1 & Weighted mean attention distance at $\ell = k^*$ \\
\texttt{attn\_span\_final} & 1 & Same at $\ell = L-1$ \\
\texttt{local\_mass\_stable} & 1 & Attention mass in local window $[t_q-5, t_q)$ at $k^*$ \\
\texttt{local\_mass\_final} & 1 & Same at $\ell = L-1$ \\
\bottomrule
\end{tabular}
\end{center}

These features encode \textit{relative} geometry (near vs.\ far) without absolute position indices.

\subsection{Detailed Component Specifications}
\label{appendix:component-details}

\subsubsection{Component A: Trajectory Features}

\paragraph{Layer-wise probability.}
\[
p^{(\ell)}_t =
\begin{cases}
\text{softmax}(\mathbf{Z}^{(\ell)}[t])[\hat{y}^{(L-1)}_t] & \text{if } \hat{y}^{(\ell)}_t = \hat{y}^{(L-1)}_t \\
0 & \text{otherwise}
\end{cases}
\]

We record probability only when the layer's prediction matches the final prediction. This focuses the trajectory on confidence \textit{in the converged answer}.

\paragraph{Margin over second-best.}
\[
m^{(\ell)}_t = p^{(\ell)}_t - p^{(\ell)}_{t,2}
\]
where $p^{(\ell)}_{t,2}$ is the probability of the second-highest token at layer $\ell$. Margin distinguishes confident-deterministic (high margin) from confident-competitive (low margin) predictions.

\paragraph{Confidence drops.}
\[
\delta^{(\ell)}_t = \max(0, p^{(\ell-1)}_t - p^{(\ell)}_t)
\]

Non-zero drops indicate self-correction: the model reconsidered its prediction between layers.

\subsubsection{Component B: Stability Metrics}

\paragraph{Stability layer.}
\[
k^* = \min \left\{ k : \forall \ell \in [k, L-1], \; \hat{y}^{(\ell)}_t = \hat{y}^{(L-1)}_t \right\}
\]

The earliest layer at which the prediction matches the final prediction and remains stable. If no such layer exists, $k^* = L$ indicates ``never stabilized.''

\paragraph{Stability continuity.}
\[
\kappa = \max \left\{ j - i + 1 : \hat{y}^{(\ell)}_t = y_t \text{ for all } \ell \in \{i, \ldots, j\} \right\}
\]

The maximum consecutive layers predicting the ground-truth token. This captures oscillation patterns: $\kappa = L$ indicates stable from first correct prediction; $\kappa = 1$ indicates only the final layer was correct.

\subsubsection{Component C: Head Activation}

For each head $(layer, head)$ at index $i = layer \cdot H + head$:

\[
\texttt{head\_act}[i] = \max_{t_k < t_q} \tilde{A}^{(layer)}_{head}[t_q, t_k]
\]

This is the \textbf{peak gated attention} for the head---the maximum attention weight to any previous position. The $\max$ operation achieves permutation invariance by discarding which position achieved the maximum.

\textbf{Stored at two layers:}
\begin{itemize}[nosep]
\item \texttt{head\_act\_stable}: Values at stability layer $k^*$
\item \texttt{head\_act\_final}: Values at final layer $L-1$
\end{itemize}

\textbf{Storage layout:} For stable layer $k^*$, values are stored at indices $[k^* \cdot H, (k^*+1) \cdot H)$. Other layer indices are zero.

\subsubsection{Component D: Head Entropy}

For each head, entropy measures attention sharpness:

\[
H(\tilde{A}^{(\ell)}_{h}) = -\sum_{t_k < t_q} \frac{\tilde{A}^{(\ell)}_{h}[t_q, t_k]}{\sum_{t_k'} \tilde{A}^{(\ell)}_{h}[t_q, t_k']} \log \frac{\tilde{A}^{(\ell)}_{h}[t_q, t_k]}{\sum_{t_k'} \tilde{A}^{(\ell)}_{h}[t_q, t_k']}
\]

Note the renormalization: we compute entropy of the \textit{normalized} distribution, not the gated weights directly. This measures pattern sharpness independent of gate magnitude.

\paragraph{Ghost head entropy gating.}
If $\texttt{head\_act}[i] < 0.1$, we set $\texttt{head\_ent}[i] = 0$. Renormalizing noise in suppressed heads amplifies random fluctuations into artificial variance that confuses clustering.

\subsubsection{Component E: Position Features (Optional)}

\paragraph{Attention span.}
\[
\texttt{attn\_span}^{(\ell)} = \frac{\sum_h \sum_{t_k < t_q} \tilde{A}^{(\ell)}_h[t_q, t_k] \cdot (t_q - t_k)}{\sum_h \sum_{t_k < t_q} \tilde{A}^{(\ell)}_h[t_q, t_k]}
\]

Weighted average distance from query position to attended positions.

\paragraph{Local mass.}
\[
\texttt{local\_mass}^{(\ell)} = \sum_h \sum_{t_k = t_q - w}^{t_q - 1} \tilde{A}^{(\ell)}_h[t_q, t_k]
\]

Total gated attention in a local window of size $w=5$.

\subsection{Implementation Reference}
\label{appendix:implementation}

The feature extraction is implemented in:
\begin{verbatim}
src/activation_clustering/features/feature_extractor.py
\end{verbatim}

Key implementation details:

\begin{itemize}[nosep]
\item \textbf{Query position}: For predicting token at position $t$, the query is at $t_q = t - 1$ (the previous token attends to context to predict the next token).

\item \textbf{Attention slicing}: \texttt{attn\_slice = layer\_attn[:, query\_pos, :query\_pos]} extracts attention from query to all previous positions.

\item \textbf{Ground truth handling}: If ground truth is not provided, the final-layer prediction is used as the target for trajectory computation.

\item \textbf{Layer logits}: Requires per-layer logits from a Per-Layer Supervision architecture or logit lens projection.
\end{itemize}

\subsection{Feature Vector Layout}
\label{appendix:layout}

For Tier 2 with $L=6$, $H=6$ (163 dimensions):

\begin{center}
\begin{tabular}{@{}llll@{}}
\toprule
Component & Indices & Size & Content \\
\midrule
\texttt{traj\_prob} & 0--5 & 6 & Layer probabilities \\
\texttt{traj\_margin} & 6--11 & 6 & Layer margins \\
\texttt{traj\_drops} & 12--16 & 5 & Inter-layer drops \\
$k^*$ & 17 & 1 & Stability layer \\
$\kappa$ & 18 & 1 & Stability continuity \\
\texttt{head\_act\_stable} & 19--54 & 36 & Peak attention at stable layer \\
\texttt{head\_act\_final} & 55--90 & 36 & Peak attention at final layer \\
\texttt{head\_ent\_stable} & 91--126 & 36 & Entropy at stable layer \\
\texttt{head\_ent\_final} & 127--162 & 36 & Entropy at final layer \\
\bottomrule
\end{tabular}
\end{center}

With optional position features (+4 dimensions):
\begin{center}
\begin{tabular}{@{}llll@{}}
\toprule
Feature & Index & Size & Content \\
\midrule
\texttt{attn\_span\_stable} & 163 & 1 & Attention span at stable layer \\
\texttt{attn\_span\_final} & 164 & 1 & Attention span at final layer \\
\texttt{local\_mass\_stable} & 165 & 1 & Local mass at stable layer \\
\texttt{local\_mass\_final} & 166 & 1 & Local mass at final layer \\
\bottomrule
\end{tabular}
\end{center}

\subsection{Validation Against Raw Activations}
\label{appendix:raw-baseline}

To confirm that engineered features capture computational structure rather than vocabulary, we cluster raw layer-5 activations (768D) and compute the Adjusted Rand Index against our feature-based clustering:

\[
\text{ARI}(\text{engineered}, \text{raw}) = 0.008
\]

The near-zero ARI indicates the clusterings share almost no structure. Raw clusters partition by token identity (punctuation, verbs, nouns); engineered clusters partition by processing strategy (early vs.\ late convergence, high vs.\ low entity attention).

This validates that our features capture structure \textit{orthogonal} to vocabulary---the fundamental requirement for discovering computational modes rather than semantic categories.

%% file: sections/appendix_topk_clustering.tex

\section{Sorted Top-K Attention Features}
\label{appendix:topk}

This appendix documents an extended feature representation that captures richer attention structure through sorted top-$k$ attention values.

\subsection{Motivation}

The base feature set (Appendix~\ref{appendix:feature-spec}) uses max-pooling to extract a single scalar per attention head:
\[
a_h = \max_{t_k} \tilde{A}_h[t_q, t_k]
\]

While permutation-invariant, this discards distributional information. A head attending strongly to one position ($[0.9, 0.05, 0.05]$) produces the same feature as one with diffuse attention ($[0.9, 0.08, 0.02]$), despite different computational signatures.

\subsection{Sorted Top-K Representation}

We extend max-pooling to retain the $k$ largest attention values per head, sorted in descending order:
\[
\mathbf{a}_h^{(k)} = \text{sort}_\downarrow\left( \tilde{A}_h[t_q, :t_q] \right)_{1:k} = (a_h^{(1)}, a_h^{(2)}, \ldots, a_h^{(k)})
\]

This representation:
\begin{itemize}[nosep]
\item \textbf{Subsumes max-pooling}: $a_h^{(1)} = \max_{t_k} \tilde{A}_h[t_q, t_k]$
\item \textbf{Preserves permutation invariance}: Sorting discards position information
\item \textbf{Captures attention shape}: Distinguishes sharp vs.\ diffuse distributions
\item \textbf{Enables distributional analysis}: The decay pattern $a_h^{(1)} \to a_h^{(k)}$ characterizes head behavior
\end{itemize}

\paragraph{Ghost head handling.}
For suppressed heads (gate $g_h < \theta$), we set $\mathbf{a}_h^{(k)} = \mathbf{0}$ rather than recording noise from inactive attention patterns. This is equivalent to $\text{sort}([0, 0, \ldots, 0])$.

\subsection{Feature Vector Structure}

With $k=5$, the extended feature vector has 455 dimensions:

\begin{center}
\small
\begin{tabular}{@{}llr@{}}
\toprule
Component & Description & Dims \\
\midrule
A. Trajectory & Layer-wise $p^{(\ell)}$, margin, drops & 17 \\
B. Stability & $k^*$, $\kappa$ & 2 \\
C. Head Top-K (stable) & $\mathbf{a}_h^{(k)}$ at layer $k^*$ & $L \times H \times k = 180$ \\
D. Head Top-K (final) & $\mathbf{a}_h^{(k)}$ at layer $L-1$ & $L \times H \times k = 180$ \\
E. Head Entropy & $H(\tilde{A}_h)$ at stable and final & $2 \times L \times H = 72$ \\
F. Position & Attention span, local mass & 4 \\
\midrule
\multicolumn{2}{@{}l}{Total} & 455 \\
\bottomrule
\end{tabular}
\end{center}

\subsection{Clustering Methodology}

High-dimensional features ($D=455$) challenge density-based clustering. We employ a two-stage approach:

\paragraph{Stage 1: Dimensionality reduction.}
PCA projects standardized features to 30 dimensions, retaining 73.9\% of variance. This balances dimensionality reduction against information loss.

\paragraph{Stage 2: Density-based clustering.}
HDBSCAN identifies clusters as dense regions in the reduced space:

\begin{center}
\small
\begin{tabular}{@{}ll@{}}
\toprule
Parameter & Value \\
\midrule
\texttt{min\_cluster\_size} & 10 \\
\texttt{min\_samples} & 1 \\
PCA components & 30 \\
UMAP \texttt{n\_neighbors} & 15 \\
UMAP \texttt{min\_dist} & 0.1 \\
\bottomrule
\end{tabular}
\end{center}

The aggressive parameters (\texttt{min\_cluster\_size}=10, \texttt{min\_samples}=1) maximize cluster discovery at the cost of some noise sensitivity.

\subsection{Results}

Applied to 50,453 tokens from the grade-school corpus:

\begin{itemize}[nosep]
\item \textbf{555 clusters} identified
\item \textbf{46.0\% coverage} (54.0\% unassigned as transitional)
\item \textbf{Clean depth separation}: Most clusters have $\sigma(k^*) \approx 0$, indicating homogeneous processing depth within clusters
\end{itemize}

\paragraph{Representative clusters.}
Table~\ref{tab:topk-clusters} shows example clusters spanning the stability depth spectrum. Clusters organize primarily by processing depth, with distinct computational signatures at each level.

\begin{table}[h]
\centering
\small
\caption{Example clusters from sorted top-5 feature clustering, organized by stability depth $k^*$.}
\label{tab:topk-clusters}
\begin{tabular}{@{}clrl@{}}
\toprule
Depth & Cluster & Size & Representative Tokens \\
\midrule
0 & C316 & 292 & collocations: ``continued[,]'', ``to[find]'' \\
0 & C399 & 292 & dialogue markers, fixed phrases \\
\midrule
1 & C98 & 558 & connectors: ``But'', ``South'', ``here'' \\
\midrule
2 & C72 & 433 & semantic predicates: ``sees'', ``safe'' \\
2 & C136 & 281 & completions: ``.'', ``not'', ``knew'' \\
\midrule
6 & C547 & 4953 & generic context-dependent \\
6 & C528 & 1121 & discourse: ``feel'', ``While'', ``as'' \\
6 & C125 & 500 & complex: ``be'', subwords (``oney'') \\
6 & C155 & 469 & technical: ``plates'', ``Palestine'' \\
\bottomrule
\end{tabular}
\end{table}

\paragraph{Interpretation.}
Depth-0 clusters capture immediately recognizable patterns from local context: collocations (``to find''), dialogue punctuation (``continued,''), and fixed phrases. Depth-1/2 clusters require moderate context integration for semantic predicates and discourse markers. Depth-6 clusters contain context-dependent tokens where prediction requires full-depth processing---the same surface form (``the'', ``a'') routes through different attention patterns depending on discourse context and domain-specific terminology.

The sorted top-$k$ features enable finer discrimination than max-pooling alone, distinguishing tokens that converge at the same depth but via different attention distributions. The largest cluster (C547, $n=4953$) has high mean distance, suggesting it serves as a catch-all for heterogeneous depth-6 patterns not captured by more specific clusters.

\subsection{Future Directions}

Two additional extensions remain unexplored:

\paragraph{Prediction-weighted attention.}
Weight attention by prediction contribution:
$\tilde{a}_{(\ell,h)}^{(t_k)} = \tilde{A}^{(\ell)}_h[t_q, t_k] \cdot p^{(\ell)}(\hat{y} \mid t_k)$.
Sorting these values maintains permutation invariance while encoding the attention-prediction relationship.

\paragraph{Cross-head order statistics.}
Rather than order statistics within each head, compute them across heads within a layer:
$\mathbf{g}^{(\ell)} = \text{sort}_\downarrow(\{a^{\max}_{(\ell,h)} : h \in [H]\})$.
This captures the distribution of head activations without encoding which specific heads achieved each rank.

\begin{figure}[t]
\centering
\includegraphics[width=0.9\columnwidth]{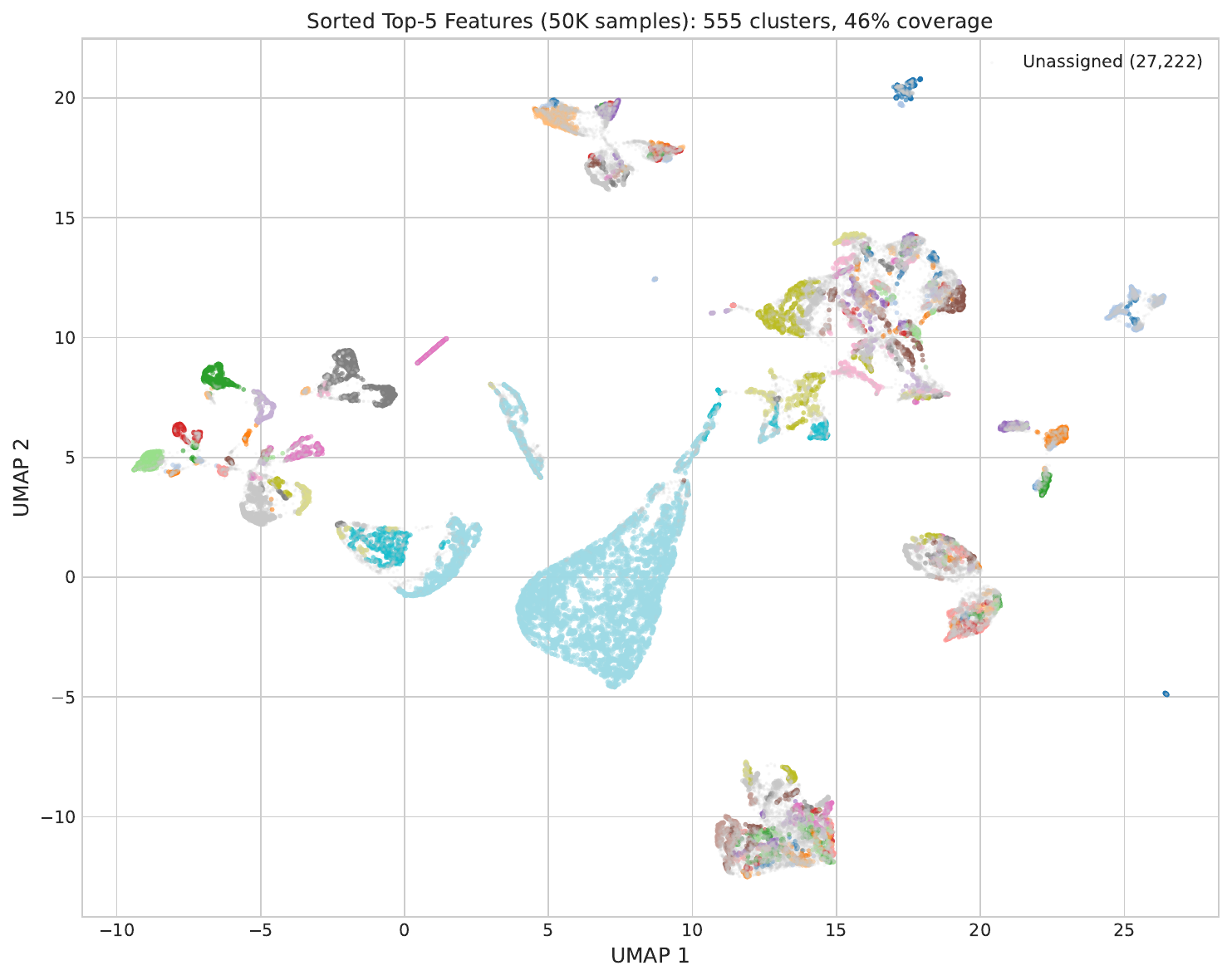}
\caption{\textbf{UMAP projection of sorted top-5 features.} 555 clusters identified by HDBSCAN (colored), with unassigned points in gray (50,453 tokens). Clusters exhibit clear spatial separation corresponding to distinct computational modes.}
\label{fig:topk-umap}
\end{figure}

\begin{figure}[t]
\centering
\includegraphics[width=0.9\columnwidth]{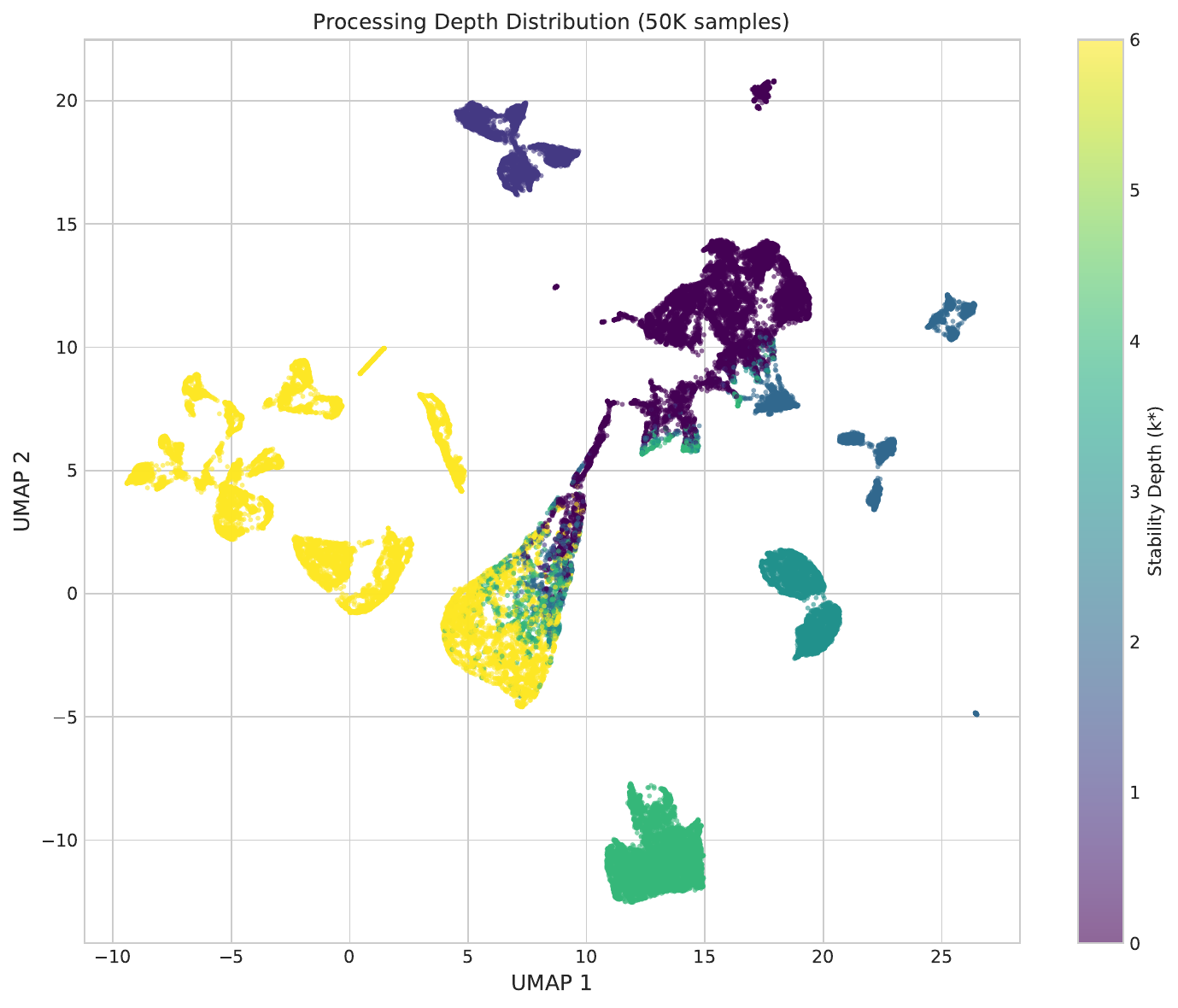}
\caption{\textbf{Processing depth in feature space.} The same UMAP projection colored by stability depth $k^*$. Spatial regions correspond to depth strata, with early-converging tokens (yellow, $k^*=0$) separated from late-converging tokens (purple, $k^*=6$).}
\label{fig:topk-depth}
\end{figure}